\documentclass[3p,review,12pt]{elsarticle}

\usepackage{hyperref,url}
\usepackage{lineno}
\usepackage{booktabs}
\usepackage{tcolorbox}
\usepackage{amsmath}
\usepackage{amsfonts}
\usepackage{stfloats}
\usepackage{adjustbox}
\usepackage{algorithmic}
\usepackage{algorithm}
\usepackage{booktabs}
\usepackage{multirow}
\usepackage{mathtools}
\usepackage{multirow}
\usepackage{verbatim}
\usepackage{amsmath,amssymb}
\usepackage{multirow,multicol}
\usepackage{tikz}
\usepackage{lineno,hyperref}
\usepackage{subcaption,graphicx}
\usepackage{hyperref}
\usepackage{devanagari}
\usepackage{xcolor}
\usepackage{algorithm,algorithmic}
\usepackage{lscape}
\usepackage{tfrupee}
\modulolinenumbers[5]
\begin{document}
\begin{frontmatter}

\title{Joint Language Identification of Code-Switching Speech using Attention based E2E Network}

\author{Ganji Sreeram}
\ead{s.ganji@iitg.ernet.in}
\author{Kunal Dhawan}
\ead{k.dhawan@iitg.ernet.in}
\author{Kumar Priyadarshi}
\ead{k.priyadarshi}
\author{Rohit Sinha}
\ead{rsinha@iitg.ernet.in}
\address{Department of Electronics and Electrical Engineering, \\Indian Institute of Technology Guwahati, Guwahati-781039, India.}

\begin{abstract}
Language identification (LID) has relevance in many speech processing applications. For the automatic recognition of code-switching speech, the conventional approaches often employ an LID system for detecting the languages present within an utterance. In the existing works, the LID on code-switching speech involves modelling of the underlying languages separately. In this work, we propose a joint modelling based LID system for code-switching speech. To achieve the same, an attention-based end-to-end (E2E) network has been explored. For the development and evaluation of the proposed approach, a recently created Hindi-English code-switching corpus has been used. For the contrast purpose, an LID system employing the connectionist temporal classification-based  E2E network is also developed. On comparing both the LID systems, the attention based approach is noted to result in better LID accuracy. The effective location of code-switching boundaries within the utterance by the proposed approach has been demonstrated by plotting the attention weights of E2E network.\\
\end{abstract}

\end{frontmatter}

\section{Introduction}
\label{sec:intro}

The phenomenon of switching between two or more languages while speaking in multilingual communities is referred to as the code-switching~\cite{Gumperz_1982_Discourse, eastman1992, Myers_1992_Comparing}. It occurs not only in verbal discourse but also in textual chats on social media platforms~\cite{bali2014, Das_2015_Code}. In a typical code-switching sentence, the syntactical composition belongs to one language while the words from the other language are used for emphasis or ease of delivery~\cite{malik1994socio, su2001code}. The syntax providing language is referred to as the \emph{native} language while the other language is called as the \emph{foreign} language. Code-switching has become a common practice in several multilingual communities across the world. The  salient examples of those include:  Arabic-English~\cite{hamed2017building}, French-Arabic~\cite{amazouz2018french}, French-German~\cite{imseng2012mediaparl}, Frisian-Dutch~\cite{yilmaz2016longitudinal}, Hindi-English~\cite{malhotra1980hindi, dey2014hindi}, Malay-English~\cite{ahmed2012automatic}, Mandarin-English~\cite{lyu2010seame}, Mandarin-Taiwanese~\cite{Lyu_2006_Speech}, and Spanish-English~\cite{solorio2008}. Code-switching poses some interesting research challenges to speech recognition~\cite{Lyu_2006_Speech, Bhuvanagirir_2012_Mixed}, language identification~\cite{Lyu_2008_Language}, language modelling~\cite{Cao_2010_Semantics, Yeh_2010_Integrated} and speech synthesis~\cite{sitaram2016speech}. However, researchers working in code-switching domain are often constrained by the lack of domain specific resources. Towards addressing that constraint, we have recently created a Hindi-English code-switching text and speech corpora referred to as the {\emph {HingCoS Corpus}}\footnote{\emph{www.iitg.ac.in/eee/emstlab/HingCoS\_Database/HingCoS.html}}. An initial version of the work describing the said corpus is available at~\cite{hingcos_2018}.

%

The task of detecting the languages present in spoken or written data using machines is referred to as language identification (LID). It finds applications in many areas including automatic recognition of code-switching speech. In~\cite{Lyu_2008_Language}, the authors developed an LID system for code-switching speech by employing separate large vocabulary continuous speech recognizers (LVCSRs). In this work, we aim to develop an LID system that can directly identify the code-switching instances instead of separately modelling the underlying languages. Recently, researchers have explored end-to-end (E2E) networks in many speech/text processing applications. The E2E networks can be trained by employing two techniques: (i) connectionist temporal classification (CTC)~\cite{graves2006connectionist}, and (ii) sequence to sequence modelling with attention mechanism~\cite{chorowski2014end}. Current literature amply demonstrates that the attention-based E2E systems outperform the CTC-based E2E systems. Recently, an utterance-level LID system employing attention-mechanism is explored~\cite{geng2016end}. In that, for producing attentional vectors, a set of pretrained language category embeddings are used as a look-up table.  Motivated by those works, we develop a joint LID system for code-switching speech using an attention-based E2E network. Unlike~\cite{geng2016end}, the attention provided for the LID system is intra-sentential~\cite{zirker2007intrasentential} and is dynamic. The salient contributions of this work include: (i) a novel application of E2E networks in developing a join LID system for code-switching speech, and (ii) demonstration of the effectiveness of the attention mechanism in locating the code-switching instances.

The remainder of this paper is organized as follows: In  Section~\ref{sec:proposed}, the proposed joint LID system trained by employing E2E networks has been discussed in detail. The detailed description of the HingCoS corpus, the system tuning parameters and the evaluations metrics involved in this study are described in Section~\ref{sec:exp}. The evaluation results along with a brief discussion, followed by the demonstration of attention mechanism for the LID task has been presented in Section~\ref{sec:results}. Finally, the paper is concluded in Section~\ref{sec:conclusion}.

\section{E2E Network based Joint LID System}
\label{sec:proposed}
In this section, we describe the creation of E2E LID systems for identifying the languages present in code-switching speech data. The developed systems jointly model the underlying languages and can handle intra-sentential code-switching types. So, in this work, we refer to them as \emph{joint LID systems}.  For developing those systems, we first explore the CTC-based E2E network and it is followed by experimentation on listen-attend-spell (LAS)~\cite{chan2016listen} network which employs an attention mechanism. The details of the said two approaches are briefly discussed below. 

\subsection{CTC-based E2E Network}
\begin{figure}[t] 
     \centering
     \hspace*{-2mm}\includegraphics[width=8cm]{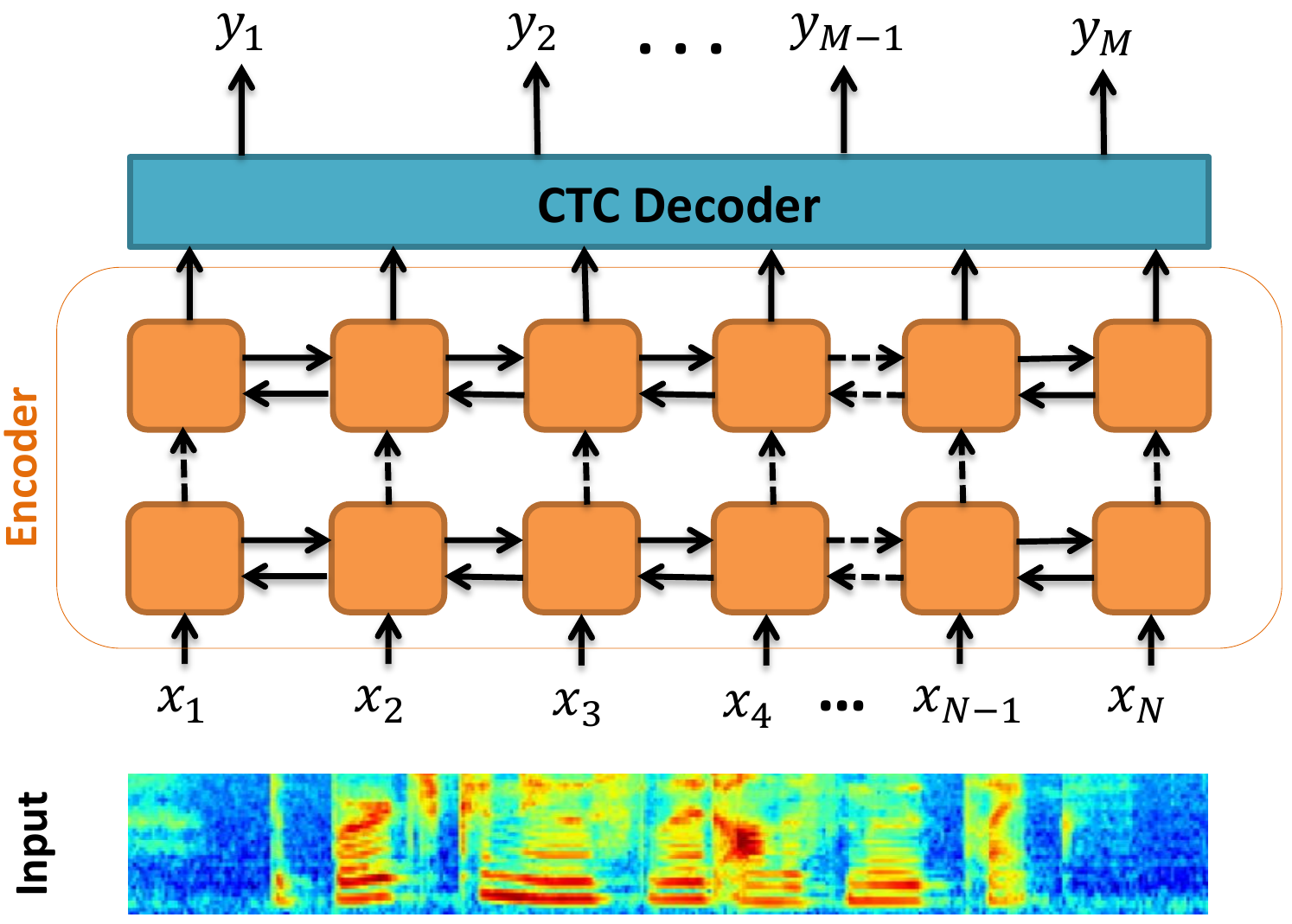}
     \caption{Architecture of the CTC-based E2E network. The encoder is a deep network consisting of BiLSTMs.}\vspace{-2mm}
     \label{fig:ctc}
\end{figure}
The CTC-based E2E architecture consists of two modules: a deep bidirectional long-short-term-memory (BiLSTM) network as an encoder, and a CTC decoder. The deep BiLSTM network encodes input feature vector ${\boldsymbol x}$ into a higher level representation vector. CTC enables the training of E2E models without requiring a prior alignment between input and output sequences.  It assumes the outputs at different time steps to be conditionally independent. The CTC decoder outputs a probability distribution over all possible output labels ${\boldsymbol y}$, conditioned on a given input sequence ${\boldsymbol x}$. A dynamic programming based forward-backward algorithm is employed to obtain the sum over all possible alignments and produces the probability of output sequence given a speech input. The typical architecture of the CTC-based E2E network is shown in Figure~\ref{fig:ctc}. Given a target transcription ${\boldsymbol y}$ and the input feature vector ${\boldsymbol x}$, the network is trained to minimize the CTC cost function as
\begin{equation}
\text{CTC}(\vec{x}) = - \text{log P}({\boldsymbol y}|{\boldsymbol x}) \nonumber
\end{equation}
$\text{where }\text{P}(\vec{y}|\vec{x}) = \sum_{\vec{a}\in \vec \beta(\vec{y,~x})}{} \text{P}(\vec{a}|\vec{x})$, $a$ is an alignment, and $\beta(\vec{y,~x})$ is the set of all possible sequences between $\vec{y}$ and $\vec{x}$. 

 \begin{figure}[!t] 
      \centering
      \hspace{-2mm}\includegraphics[width=8.1cm]{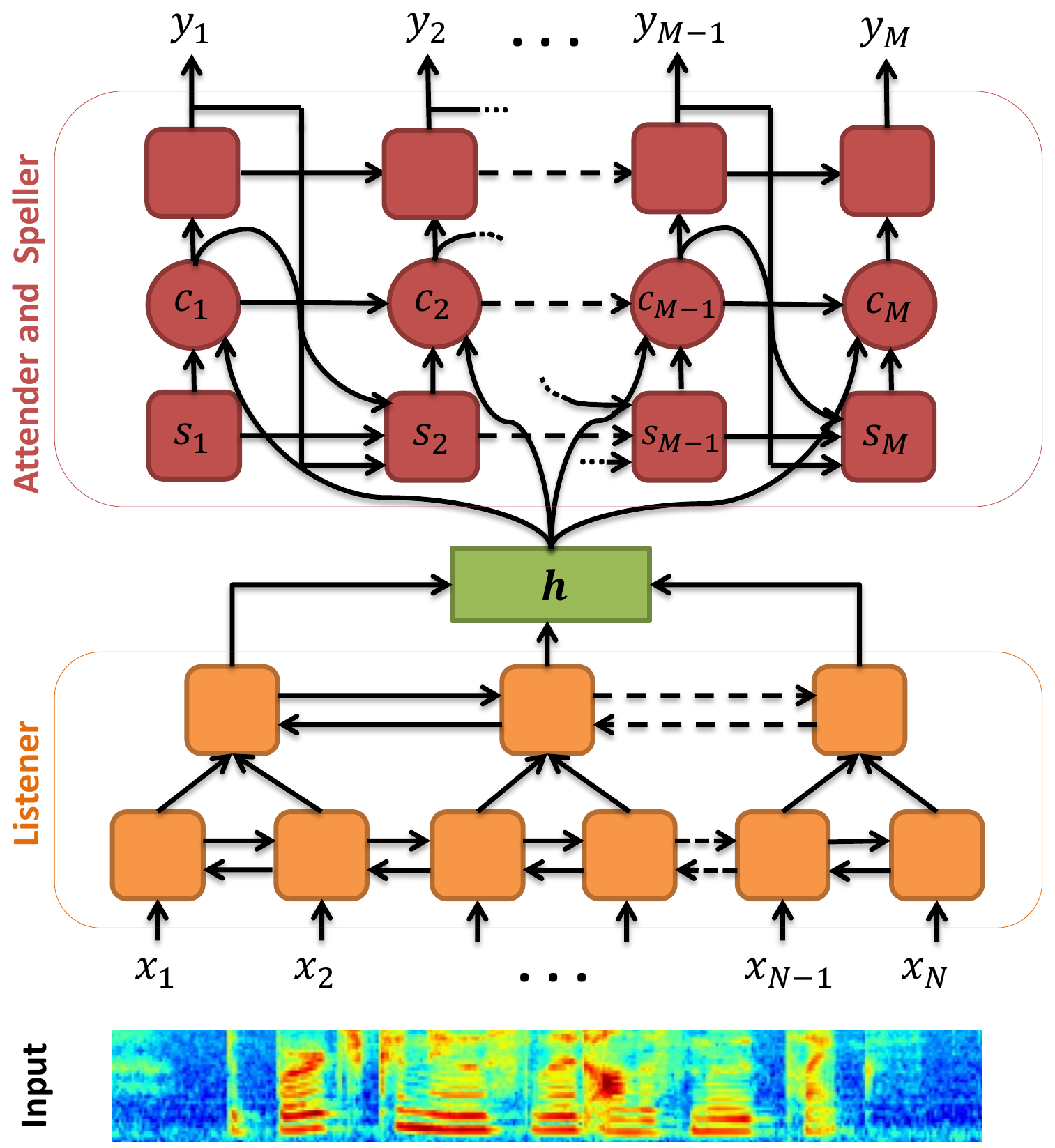}
      \caption{Architecture of LAS network. It consists of three modules namely: listener (encoder), attender (alignment generator), and speller (decoder).}\vspace{-2mm}
      \label{fig:las}
   \end{figure}
\subsection{Attention-based E2E Network}

The LAS architecture comprises of 3 modules:  listener,  attender, and speller. The listener is a pyramidal architecture consisting of BiLSTM cells. It acts as an encoder and transforms an input feature vector ${\boldsymbol x}$ into a higher order vector representation ${\boldsymbol h}$. The encoded output vector ${\boldsymbol h}$ along with the decoder state $s_{i}$ is passed to the attender. At every time instance, the attender takes ${\boldsymbol h}$ and decoder state $s_{i}$ as the inputs and outputs the context $c_{i}$. It acts like an alignment generator determining which encoded features in ${\boldsymbol h}$ should be attended for accurate prediction of the current output symbol $y_i$. The output of this attention module $c_i$ is then passed to the speller, which is an LSTM decoder. It takes the context information $c_i$ as well as the previous prediction $y_{i-1}$ in order to predict the current symbol $y_i$. The listener, attender and the speller are trained together to minimize the cross-entropy loss and thus making it a complete end-to-end system. The typical architecture of the LAS network is shown in Figure~\ref{fig:las}. The mathematical representations of each step in the LAS architecture are given as

\begin{eqnarray}  \nonumber
{\boldsymbol h} &=& \text{Listener}({\boldsymbol x})\\ \nonumber
c_{i }&=& \text{Attender}({\boldsymbol h}, s_{i})\\ \nonumber
s_{i }&=& \text{LSTM}(y_{i-1}, s_{i-1}, c_{i-1})\\ \nonumber
p({y_{i}}|{\boldsymbol x}) &=& \text{Speller}(c_{i}, y_{i-1}).
\end{eqnarray}  
\begin{figure*}[] 
     \centering
     \includegraphics[width=16.9cm]{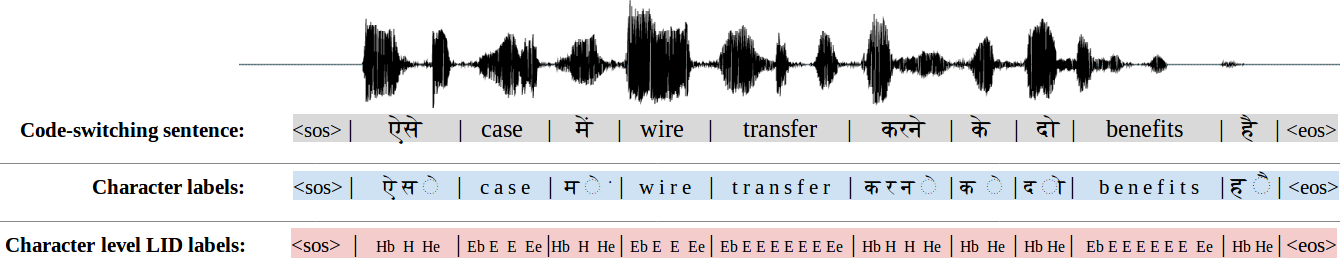}
     \caption{Creation of character-level LID tags for the training data towards conditioning the E2E networks to perform LID task on code-switching speech. The `$H/E$' denotes Hindi/English LID tag. The `$b/e$' label is appended to the `$H/E$' LID tag to mark the begin/end characters.}
     \label{fig:las_lid}
  \end{figure*}
\subsection{Creation of Target Labels for LID}
In intra-sentential code-switching utterances, the duration of the embedded foreign words/phrases could be very short. Thus, LID is required to be performed at the word level rather than the utterance level. Further, the explored E2E networks are required to be conditioned to perform the LID task on code-switching speech data.
For achieving that, for each of the training utterance, first the given orthographic transcription is transformed into character level transcription. Later, each character in the transcription is mapped to the corresponding LID tags. This process is illustrated in Figure~\ref{fig:las_lid}. This is to highlight that, in the orthographic transcription of the training data, the Hindi and English words are written in their respective scripts. So, the character-level LID tags as `$H/E$' are produced in a straight forward manner, except that additional labels `$b$' and `$e$' are appended to the tags of \emph{begin} and \emph{end} characters of each word, respectively. Also, a blank symbol `$|$' has been inserted between words to ease the marking of the word boundary. For training the E2E models, a total of $8$ labels which include $6$ LID tags ($Hb$, $H$, $He$, $Eb$, $E$, $Ee$), one blank label ($|$), and a silence label ($sil$) are given as targets to generate the output posterior probabilities. With the proposed target labelling scheme, the attention-based E2E system is hypothesized to predict the language boundaries more accurately. The experimental results discussed later in Section~\ref{sec:results} support the same.  

\section{Experimental Setup}
\label{sec:exp}
This section describes the code-switching corpus used for experimentation purposes. The description of the tuning of parameters of different LID systems developed and the metric employed for evaluating their performances are also presented. 
\subsection{Database and Front-end Features}
\label{sec:data_prep}
The HingCoS speech corpus, used for the experimentation, is collected over the telephone by speakers uttering predefined Hindi-English code-switching sentences in varying acoustic environments. These sentences were collected by crawling a few web-blogs having different contexts~\cite{hingcos_2018}. In this corpus, the native language of the sentences is Hindi and the foreign language is English. The speech data is contributed by $101$ speakers ($64$ male and $37$ female) and recorded at a sampling frequency of $8$ kHz and a resolution of $128$ bits. This database contains $9251$ Hindi-English code-switching utterances (about $25$ hours) with utterance duration varying between $2$ to $30$ seconds. For experimental purpose, the corpus is partitioned into train, test, and development sets containing $7015$, $100$ and $2136$ sentences, respectively. 

The acoustic features comprise of $26$-dimensional log filter bank energies computed using the  Hamming window having the length as $25$ ms, window shift of $10$ ms, and pre-emphasis factor of $0.97$. These features are then used to develop both CTC- and attention-based E2E LID systems. The development of these systems has been done on the Nabu toolkit~\cite{nabu_2017}.

\subsection{Parameter Tuning}
In this section, we describe the tuning of parameters for both the developed LID systems done on the development set defined earlier.  In this work, the attention-based E2E LID system is trained by employing the LAS network in which the encoder (listener)  has $2$ hidden layers, each with $128$ BiLSTM nodes. The dropout rate of the encoder is set as $0.5$. The number of hidden layers and nodes of the decoder (speller) are kept same as that of the encoder, except that the nodes are simple LSTMs. The LAS network is trained by setting the number of epochs as $300$, batch size as $32$, and the learning rate decay as $0.1$. During decoding, the beam width is set as $8$. 
For contrast purpose, a CTC-based E2E LID model is trained with $3$ hidden layers, each having $128$ BiLSTM nodes. The dropout rate of the encoder is set to be similar to that of the LAS model. The network is trained with number of epochs as $300$. Also, the  parameters corresponding to the batch size, and the learning rate decay are set as $8$ and $0.1$, respectively. During decoding, the CTC cost function is employed to produce $1$-best output sequence.
\begin{figure*}[] 
\centering
      \hspace*{-1mm}\includegraphics[scale=0.69]{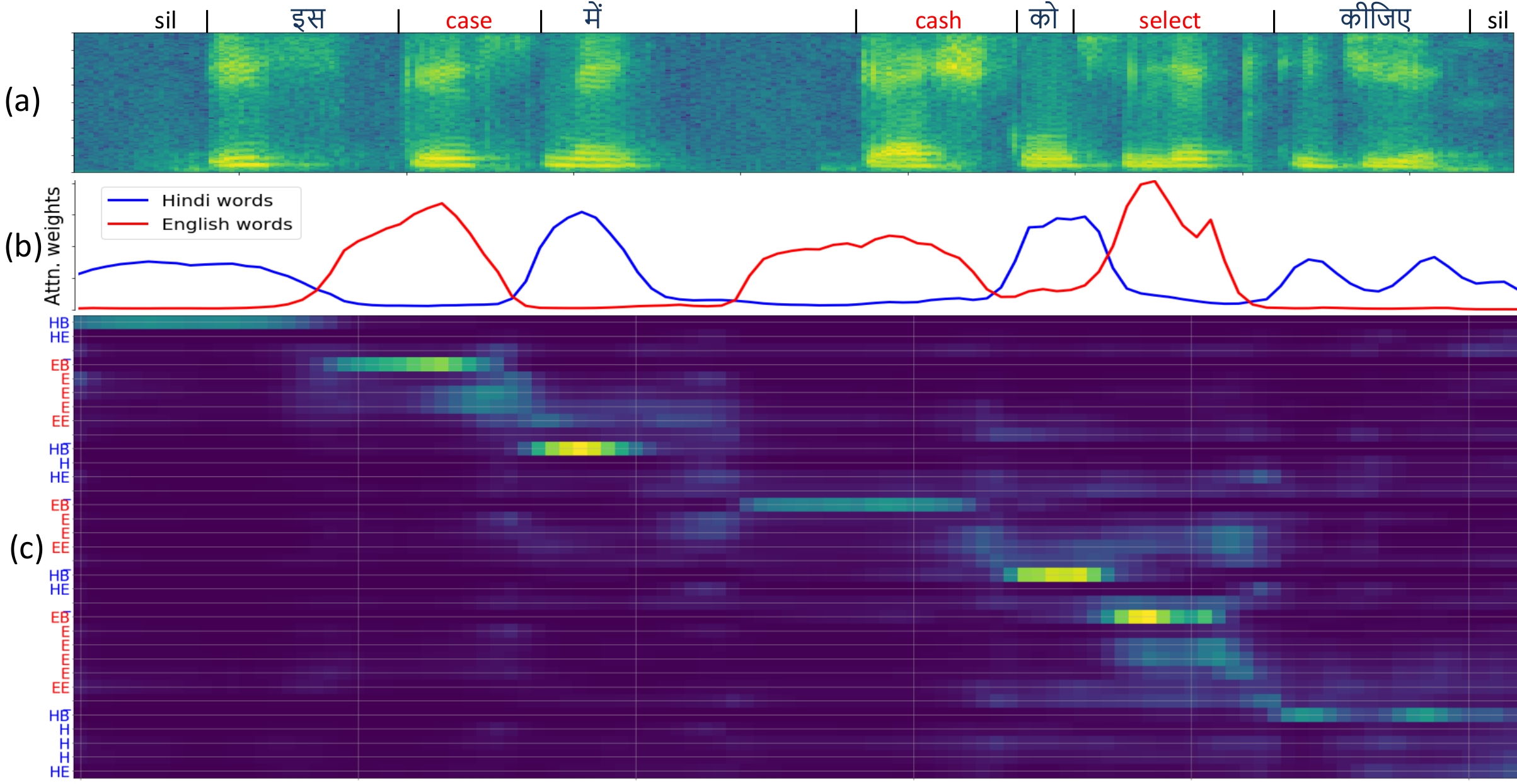}
       \caption{Visualization of attention mechanism for LID task. For a given Hindi-English code-switching utterance: a) spectrogram labeled with Hindi and English word boundaries for reference purpose. (b) variation of attention weights with respect to time for Hindi and English languages, and (c) alignment produced by the attention network for the input speech and the decoded output LID labels.}
       \label{fig:las_an}
\end{figure*}

\subsection{Evaluation Measures}
The developed E2E systems are evaluated in terms of the LID error rate computed as
\begin{equation}
\text {LID error rate} = \frac{N_S +N_I + N_D}{N}\times 100 \nonumber
\end{equation}
where, the numerator terms $N_S$, $N_I$, and $N_D$ refer to the number of substitutions, insertions, and deletions, respectively. The denominator $N$ refers to the total number of labels in the reference. For this evaluation, the reference transcriptions for all test utterances labeled in terms of the proposed LID tags are aligned with the corresponding outputs produced by the E2E network. In addition to this character-level LID error rate, a corresponding word-level LID error rate is also computed in a similar fashion by applying majority voting scheme~\cite{parhami1994voting} on the character-level LID labels.

\begin{table}[]
\centering
\caption{Evaluation of the developed E2E LID systems in Hindi-English code-switching task. The LID error rates have been computed both for character and word levels. The total number of characters/words ($N$) in the reference transcription is $198,855/41,025$.}\label{tab:results}
\scalebox{0.88}{
\begin{tabular}{|c|c|r|r|r|c|}
\hline
\textbf{\begin{tabular}[c]{@{}c@{}}LID\\ system\end{tabular}} & \textbf{\begin{tabular}[c]{@{}c@{}}Target\\ label\end{tabular}} & \multicolumn{1}{c|}{$\boldsymbol N_D$} & \multicolumn{1}{c|}{$\boldsymbol N_I$} & \multicolumn{1}{c|}{$\boldsymbol N_S$} & \textbf{\begin{tabular}[c]{@{}c@{}}LID error \\  rate (\%)\end{tabular}} \\ \hline \hline
\multirow{2}{*}{CTC}                                          & Character                                                           & 73,502                          & 3,384                           & 20,957                          & 49.20                                                                    \\ \cline{2-6} 
                                                              & Word                                                            & 3,576                           & 3,655                           & 5,136                           & 30.14                                                                    \\ \hline \hline
\multirow{2}{*}{Attention}                                    & Character                                                           & 20,299                          & 13,587                          & 12,789                          & \textbf{23.47}                                                                    \\ \cline{2-6} 
                                                              & Word                                                            & 2,713                           & 1,616                           & 2,484                           & \textbf{16.60}                                                                    \\ \hline
\end{tabular}}
\end{table}
\begin{table}[!t]
\centering
\caption{The character and word level decoded outputs for CTC- and attention-based E2E LID systems for the utterance considered in Figure~\ref{fig:las_an}. A majority voting scheme is employed for mapping the character-level LID label sequences to word-level LID label sequences. The attention-based system is able to decode the LID label sequences more accurately when compared to the CTC-based system.}\label{support}

\scalebox{0.63}{
\begin{tabular}{|c|l|l|}
\hline
\multirow{3}{*}{\textbf{\begin{tabular}[c]{@{}c@{}}Character level \\ LID lables\end{tabular}}} & {Reference sequence}                &  Hb He $|$ Eb E Ee $|$ Hb H He $|$ Eb E Ee $|$ Hb He $|$ Eb E E E E Ee $|$ Hb H H H He       \\ \cline{2-3} 
                                                                                                & {CTC-based hypothesis}       & Hb E Ee $|$ Eb E Ee $|$ Eb He $|$ Hb He $|$ Eb Ee $|$ Eb Ee $|$ Hb He                        \\ \cline{2-3} 
                                                                                                & {Attention-based hypothesis} & Hb He $|$ Eb E E E Ee $|$ Hb H He $|$ Eb E E Ee $|$ Hb He $|$ Eb E E E E Ee $|$ Hb H H H He \\ \hline \hline
\multirow{3}{*}{\textbf{\begin{tabular}[c]{@{}c@{}}Word level \\ LID lables\end{tabular}}}      & {Reference sequence}                & H $|$ E $|$ H $|$ E $|$ H $|$ E $|$ H                                                          \\ \cline{2-3} 
                                                                                                & {CTC-based hypothesis}       &  E $|$ E $|$ E $|$ H $|$ E $|$ E $|$ H                                                           \\ \cline{2-3} 
                                                                                                & {Attention-based hypothesis} & H $|$ E $|$ H $|$ E $|$ H $|$ E $|$ H                                                         \\ \hline
\end{tabular}}
\end{table}
\section{Results and Discussion}
\label{sec:results}
In this work, two different kinds of E2E joint LID systems are developed and evaluated on the HingCoS corpus. The LID error rates computed both at character and word levels for these systems are reported in Table~\ref{tab:results}. In contrast to CTC, the use of LAS architecture in E2E LID system is noted to yield substantial reduction in the error rates. This is attributed to the ability of attention mechanism in LAS network to accurately predict the languages switching in the data. To highlight that, we have computed the language-specific averaged attention weights with respect to the decoded LID label sequence and the plot for the same is shown in Figure~\ref{fig:las_an}. The description of each of the subplots in Figure~\ref{fig:las_an} is presented next. 

Figure~\ref{fig:las_an}(a) shows the spectrogram of a typical Hindi-English speech utterance in the test set. Note that, the spectrogram is manually labeled with spoken words and their boundaries for the reference purposes. The variations of the averaged attention weights for Hindi and English language targets present in the input speech data with respect to time, are shown in Figure~\ref{fig:las_an}(b). The sequence alignment produced by the attention network for the input speech data (on the x-axis) and the decoded output LID labels (on the y-axis) is plotted in Figure~\ref{fig:las_an}(c). From Figures~\ref{fig:las_an}(b) and~\ref{fig:las_an}(c), we observe that the attention weights for Hindi and English languages mostly peak around the corresponding word locations.

It is worth highlighting here that both CTC-based and attention-based E2E systems are provided with identical target-level supervision while training. Unlike the attention-based system, the CTC-based system could not exploit that supervision. This is attributed to the fact that CTC assumes the outputs at different time steps to be conditionally independent, hence making it less capable of learning the sequence. To support this argument, for the very utterance in Figure~\ref{fig:las_an}, the character level decoded outputs of the CTC- and attention-based E2E LID systems are listed Table~\ref{support}. The word-level LID labels for both the considered systems are also shown in that table. On comparing the hypothesized sequence of output labels, it can be noted that the inclusion of attention mechanism in E2E LID system leads to more effective language identification within code-switching speech data.

\section{Conclusions}
\label{sec:conclusion}
In this work, we propose joint E2E LID systems employing CTC and attention mechanism for identifying the languages present in code-switching speech. The development and evaluation of the proposed systems are done on a recently created Hindi-English code-switching speech corpus. Towards developing the LID systems, a novel target labeling scheme has been introduced which is found to be very effective for the attention-based system. On comparing the attention and CTC mechanisms, the former is noted to achieve a two-fold reduction in both character- and word-level LID error rates. The work also demonstrates the ability of the attention mechanism in detecting the language boundaries in code-switching speech data. Despite the experiments being performed on Hindi-English code-switching data, the proposed approach can easily be extended to other code-switching contexts.

In a recent work~\cite{luo2018towards}, the authors reported improvement in Mandarin-English code-switching ASR by employing multi-task learning with the LID labels. Motivated by that work, in the future, we aim to explore the proposed LID labelling scheme as a supervision in the multi-task learning framework for code-switching ASR.

\bibliography{LID}

\end{document}